\documentclass{article}

\usepackage[final]{neurips_2021}
\usepackage[pdftex]{graphicx}

\usepackage[utf8]{inputenc} % allow utf-8 input
\usepackage[T1]{fontenc}    % use 8-bit T1 fonts
\usepackage{hyperref}       % hyperlinks
\usepackage{url}            % simple URL typesetting
\usepackage{booktabs}       % professional-quality tables
\usepackage{amsfonts}       % blackboard math symbols
\usepackage{nicefrac}       % compact symbols for 1/2, etc.
\usepackage{microtype}      
\usepackage{xcolor}         % colors
\usepackage{textcomp}
\usepackage{amsfonts}
\usepackage[mathscr]{eucal}
\usepackage{graphicx}
\usepackage{mathtools}
 
\usepackage{booktabs}
\usepackage{amssymb}
\usepackage{diagbox}
\usepackage{subcaption}
\title{Communication-Efficient Federated Learning for Neural Machine Translation}

\author{%
  Tanya Roosta\thanks{Equal Contribution} \hspace{10mm} Peyman Passban$^*$ \hspace{10mm} Ankit Chadha\\
  Amazon\\
  \texttt{\{troosta, peymp, ankitrc\}@amazon.com} \\
}

\begin{document}
\maketitle
\begin{abstract}
Training neural machine translation (NMT) models in federated learning (FL) settings could be inefficient both computationally and communication-wise, due to the large size of translation engines as well as the multiple rounds of updates required to train clients and a central server. In this paper, we explore how to \textit{efficiently} build NMT models in an FL setup by proposing a novel solution. In order to reduce the communication overhead, out of all neural layers we only exchange what we term ``Controller'' layers.  Controllers are a small number of additional neural components connected to our pre-trained architectures. These new components are placed in between original layers. They act as liaisons to communicate with the central server and learn minimal information that is sufficient enough to update clients. 

We evaluated the performance of our models on five datasets from different domains to translate from \textit{German} into \textit{English}. We noted that the models equipped with Controllers preform on par with those trained in a central and non-FL setting. In addition, we observed a substantial reduction in the communication traffic of the FL pipeline, which is a direct consequence of using Controllers. Based on our experiments, Controller-based models are $\sim$\hspace{0.4mm}6 times less expensive than their other peers. This reduction is significantly important when we consider the number of parameters in large models and it becomes even more critical when such parameters need to be exchanged for multiple rounds in FL settings.  
\end{abstract}

\section{Introduction}\label{intro}
Federated learning (FL) is a paradigm in which models can be trained in a \textit{decentralized} and \textit{private} fashion. Model training in FL is distributed over multiple clients, each with their own set of data. This form of distributed training enables building models that can benefit from data residing in other clients without having direct access to them. The FL framework is different than the more traditional distributed training \citep{DBLP:journals/corr/abs-2007-03970}, where the assumption is that data on devices is \textit{accessible} and \textit{identically distributed}, referred to as the IID condition. However, this assumption does not hold in FL where each client could have a private dataset very different than others (non-IID data). This dataset heterogeneity introduces new algorithmic challenges in FL that do not exist in the distributed setting.

FL has been attracting more attention in recent years due to the inherent privacy it offers for model training. Customer privacy has been a top priority at many large companies and governing authorities, and the FL framework offers the setup for training models without the need for moving data from clients to a central server, or sharing data among clients. Despite this appealing characteristic, FL systems are still hard to implement and deploy in practice, since (in addition to the heterogeneity problem) they require an iterative communication update method. In each round, models' parameters have to be exchanged between clients and a central server.  The communication overhead of sending such a large number of parameters could become prohibitive and surpass the power budget on clients. This factor affects the practicality of FL in real-world scenarios.

Motivated by the aforementioned problem, in this paper we try to reduce the communication bandwidth.  We apply our technique to train neural machine translation (NMT) engines. The reason for this choice is that bi-lingual datasets provide a natural way to incorporate data heterogeneity at local nodes. Generally, various models are used to generate synthetic datasets that simulate this non-IID data distribution \citep{wang2021device}. However, having a truly heterogeneous dataset without the need for simulation is crucial, and helps produce realistic results. NMT and the rich dataset it offers, enables us to setup such an experimental environment organically. 

NMT also allows us to challenge the capabilities of our setup. Bilingual engines are usually large and data-hungry models. Demonstrating that such models are trainable in an FL setting has its own research values. Moreover, we leverage our novel idea to reduce communication and computation costs in the same setting. We design a structure we refer to as \textit{Controller}. A Controller is a layer inserted between two existing layers of a neural model. During training, the non-Controller (original) layers are frozen and only the Controller layer parameters are updated. Similarly, during communication, only the parameters of Controllers are shared between the clients and the central server. This is the main contribution of our research, that we will demonstrate through running experiments.

\section{Federated Learning}\label{background}
FL was introduced in  \citet{fedsurvey16} and \citet{mcmahan2017communication} as a distributed training framework to handle large scale data on edge devices, where data local to each edge device remains private \citep{9084352}. Each client is unable to generalize well given they only have access to local data with limited samples, thus limiting high-performance on practical use-cases. Therefore, clients could learn local domains but might fail to generalize to other domains. FL aims to address this as well as some other concerns through providing a platform for parties to share their model parameters. FL has been wide spread in different applications such as text analysis \citep{medicalapp}, big data \citep{bd9014272}, and computer vision \citep{cvapp}. 

FL can be implemented in various ways \citep{li2019survey}, but in this paper we use the \textit{cross-silo} form, where a server node orchestrates learning and communication among clients and maintains a central model. In each FL round, the server pulls information (gradients or parameters) from clients and combines it to produce global knowledge. In our experiments, we use {\fontfamily{pcr}\selectfont FedAVG} \citep{mcmahan2017communication} to aggregate clients' information via a simple parameter averaging technique, as shown in Equation \ref{eq1}:
\begin{equation}\label{eq1}
    w_{i} \leftarrow \sum_{m=1}^{M} \frac{n_m}{n} w^m_i
\end{equation}
where $M$ is the number of clients, $w_i$ is the server parameter set for the \textit{i}-\textit{th} training iteration, $n_m$ is the number of data points in the \textit{m}-th client's dataset, and $n$ is the total number of all training data. Besides {\fontfamily{pcr}\selectfont FedAVG}, there exist more sophisticated aggregation algorithms, such as FedOpt \citep{app10082864} and FedProx \citep{FedProx}.  Though these algorithms could lead to some performance improvement, we choose to use {\fontfamily{pcr}\selectfont FedAVG} for two reasons:
\begin{itemize}
    \item It offers an intuitive form of aggregation for FL, which provides a widely acceptable and easily reproducible baseline. This is crucial for our work since it is one of the few, if not the only, research efforts that combines FL and NMT. 
    \item It has shown good performance across many applications of FL despite its simplicity. This feature plays a crucial role when it comes to model training in real world applications. 
\end{itemize}

\subsection{Transfer Learning}\label{TL}
Traditionally, in machine learning problems, a model is trained and validated using data from a particular domain. Such a model usually does not generalize well on domains outside of its training dataset. In order to address this issue, transfer learning was proposed \citep{TransferLearning}.  The idea is to train a neural model in a particular domain, and then use the model in a different domain. Transfer learning has worked well in many fields, and the trained models in one domain can mostly generalize to other domains either straight away or with some fine-tuning.  However, when fine tuning a model in a new domain, all neural model parameters are usually updated, which might not be necessary. \citet{houlsby2019parameter}, \citet{pfeiffer2020adapterhub}, and \citet{ruckle2020adapterdrop} proposed a new set of architectures, called Adapters, to address this shortcoming. Adapters are low-cost plug-ins that are mounted on pre-trained models inside their internal layers. When fine-tuning the model, only these adapter components are updated, which makes training more parameter-efficient and less expensive.  However, adding adapter layers requires changing the internal architecture of the model, and the question of where to exactly insert these adapters to obtain performance improvement is not easily answered \citep{ruckle2020adapterdrop}.  

We argue that using FL framework to train mixed-domain NMT models (the scope of this paper) is similar to transfer learning.  Collecting parameters from clients of different domains, aggregating their information, and sending back aggregated knowledge to clients is a form of transfer learning. With this understanding, we ask \textit{how can we make this transfer learning more communication and computation efficient?} Given that adding adapter layers is not straight forward, we propose a simple yet effective \textit{Controller} alternative which we cover in detail in Section \ref{method}.

\section{Communication and Computation-Efficient Model Training in FL}
\label{method}
In FL, there is an aggregation phase in which information from multiple clients is fused on the server side, and the result is sent back to clients so they can learn from each other. This round-trip update requires a considerable communication bandwidth since \textit{all} the model parameters are exchanged in each update round. One way to limit the communication overhead is to only send a subset of parameters that are more critical and sufficient for model training. However, for such an approach we might need to discover precise selection heuristics that add to the computation overhead.

\subsection{Controller Layers}
\begin{figure*}[h]
    \centering
    \includegraphics[width=0.66\textwidth] {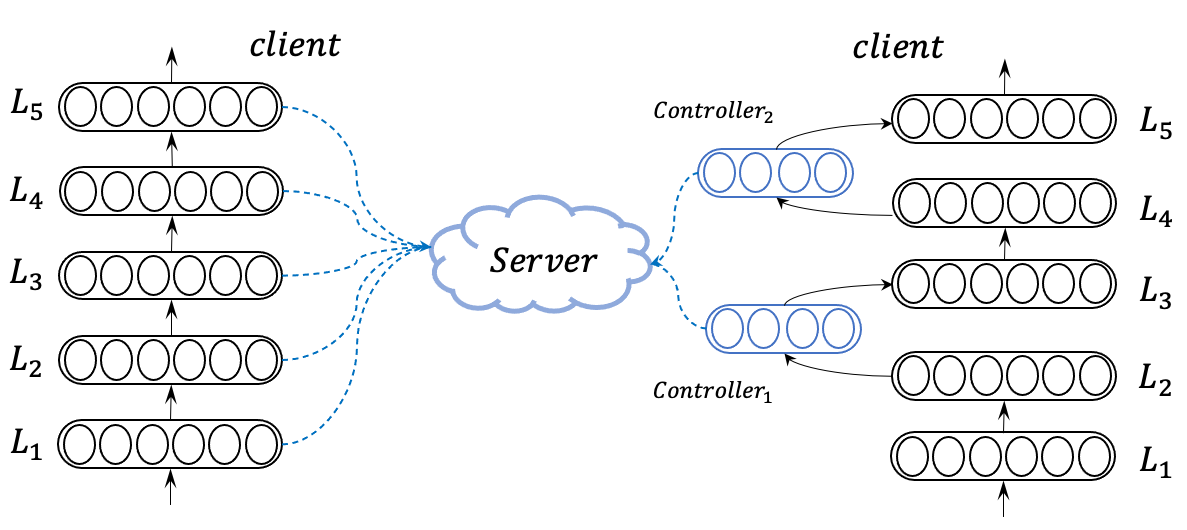}
    \caption{A comparison between a client model equipped with Controller layers and its ordinary peer.}
    \label{fig:controller}
\end{figure*}

To tackle the parameter selection problem, we propose a simple yet effective architectural addition, which we call Controllers. The Controller is another neural layer inserted in between two existing model layers. The goal of these \textit{dedicated} layers is to adapt and promote information flow from the layers \textit{below} them, and communicate that information to the central server. During the local update phase, these Controllers will disseminate the aggregate information from the server to the local models. Figure \ref{fig:controller} illustrates this concept.

In the model with Controllers, only the parameters of these particular layers are communicated with the server. Not only the communication overhead is reduced, there are additional privacy gains from this approach. Moreover, Controllers help manage the computation cost on the client side. Our Controller mechanism can be used in any neural architecture, but in our experiments, we embedded them into the Transformer \citep{attention} architecture since it is the current state-of-the-art and a well-studied model in the community, and this makes our work easy to replicate by others.

In our models, we treat the Controller layers with two strategies during training. In the first set, we update all model parameters in conjunction with the Controller layers. In the second set, we freeze the original parameters of the model and only update the Controller layers. This is to compare the performance of the model in both scenarios and understand how much we lose in performance, if we reduce the layers involved in the training process. Regardless of which option we choose, the fact that we only communicate the Controller layer parameters to the server, and only need to update them when pushing back the information to clients, helps greatly to reduce the communication overhead.  Moreover, if we freeze the original layers and only train the Controller layers, we are able to reduce the training time and computation overhead. Usually, clients in FL settings have limited resources and might not be able to train large models. Our proposed Controller layers provide benefits of an in-domain large model without incurring the computation cost on edge devices.

\section{Experimental Study}\label{exp}
We designed our experiments using five bi-lingual, German--English datasets of \textit{WMT},\footnote{\url{http://statmt.org/wmt14/translation-task.html}} \textit{OpenSubtitle} or \textit{OS} \citep{lison-tiedemann-2016-opensubtitles2016}, \textit{PHP}, \textit{Ubuntu} or \textit{UB}, and \textit{TED} \citep{TIEDEMANN12.463}. \textit{OS} is a large collection and for our experiments we only randomly selected 4.5 Million sentences of it, to make its size comparable to \textit{WMT}. Each dataset is from a different genre with a different size, which provides a heterogeneous collection of data for evaluation. For the test and development sets, we use \textit{newstest-14} and \textit{newstest-13}, respectively, for \textit{WMT}, and for the other datasets, we randomly select 2K sentences for the test and 2K for the development sets. 

All datasets are pre-processed with a \textit{BPE} model \citep{sennrich-etal-2016-neural} to extract sub-word units. Both English and German sides are used to train the \textit{
BPE} model; we extracted 30K shared sub-units for each side. As our client and server models, we use Transformers. Each model has 6 layers on the encoder side and 6 layers on the decoder side. \textit{All} settings including hyper-parameters, dimensions, and training steps are identical to those of \citet{attention}. All of our baseline and FL results are summarized in Table \ref{results}. 
\begin{table*}[th]
\small
\begin{center}
\begin{tabular}{l | c c c c c | c | c c}
\toprule
Model $\blacktriangledown$ $|$ Testset $\blacktriangleright$ & \textit{WMT}$_{ts}$ & \textit{OS}$_{ts}$ & \textit{TED}$_{ts}$ & \textit{PHP}$_{ts}$ & \textit{UB}$_{ts}$ & \textit{Average} & \textit{C-Cost} & \textit{T-Cost}\\\midrule
\multicolumn{9}{c}{Non-FL Setting}\\\hline
 \textbf{ 1}: {6L-6D WMT} & 33.66 & 18.57 & 29.22 & 8.04 & 12.41 & 20.38 & NA & 6E-6D + W\\
 \textbf{ 2}: {6E-6D OS} & 13.66 & 23.58 & 24.22 & 7.84 & 13.83 & 16.62 &  NA & 6E-6D + W\\ 
 \textbf{ 3}: {6E-6D TED} & 12.09 & 13.59 & {29.32} & 6.67 & 10.15 & 14.36 &  NA & 6E-6D + W\\
 \textbf{ 4}: {6E-6D PHP} & 0.00 & 0.26 & 0.26 & 34.48 & 0.00 & 7.00 &  NA & 6E-6D + W\\
 \textbf{ 5}: {6E-6D UB} & 0.28 & 0.78 & 0.75 & 2.30 & 30.15 & 6.85 &  NA & 6E-6D + W \\\hline
 \textbf{ 6}: Fine Tuning & 33.50 & 21.82 & 31.51 &  37.56 & 35.61 & 32.00 &  NA & 6E-6D + W\\
 \textbf{ 7}: Chained Training & 18.26 & {23.51} & 28.19 & 16.14 & 23.05 & 21.83 &  NA & 6E-6D + W\\\hline
 \multicolumn{9}{c}{FL Setting}\\\hline
 \textbf{ 8}: 6E-6D/A-A & 33.97 & 19.17 & 30.8 & 37.32 & 47.9 & 33.83 & 6E-6D & 6E-6D + W\\
 \textbf{ 9}: 8E-8D/A-A & 29.17 & 18.6 & 28.62 & 32.26 & 33.36 & 28.40 & 8E-8D & 8E-8D + W\\
 \textbf{10}: 8E-8D/A-C (2-6) & 29.03 & 18.59 & 28.44 & 31.99 & 33.75 & 28.36 & 8E-8D & 2E-2D \\
 \textbf{11}: 8E-8D/A-C (0-6) & 31.79 & 20.02 & 30.6 & 32.43 & 33.41 & 29.65 & 8E-8D & 2E-2D \\
 \textbf{12}: 8E-8D/C-C (2-6) & 33.75 & 19.51 & 30.34 & 32.29 & 31.44 & 29.46 & 2E-2D & 2E-2D\\
 \textbf{13}: 8E-8D/C-C (0-5) & 31.9 & 20.00 & 30.91 & 31.88 & 32.74 & 29.48 & 2E-2D & 2E-2D\\\hline
 \textbf{14}: 6E-6D/C-C (0-3) & 31.13 & 19.19 & 30.95 & 33.79 & 32.85 & 29.58 & 2E-2D & 2E-2D\\
 \textbf{15}: 6E-6D/C-C (1-4) &  29.51 & 18.62 & 29.5 & 31.13 & 29.61 & 27.67 & 2E-2D & 2E-2D\\
\bottomrule
\end{tabular}
\caption{\label{results} Results from our NMT engines when trained in both central and FL settings. The first column shows the model and its configuration, e.g. in Row \textbf{1} the \textit{6E-6D WMT} model is a Transformer with 6 encoder and 6 decoder layers trained using the WMT training set. The \textit{ts} subscript in other columns indicate the test sets, i.e. the score in Row \textbf{1}, Column \textit{OS}$_{ts}$ shows the performance of the \textit{6E-6D WMT} model on the \textit{OS} testset. In the FL setting, each model is labeled with a character tuple followed by a pair of digits. The first letter in the tuple indicates if all (A) or only Controller layers (C) are shared during communication. The second letter shows if we train all layers (A) or only update values of the Controller layers (C). The digit pair defines the position of the Controller layers, e.g. in the (2-6) setting in Row \textbf{13}, the first Controller layers is placed after the second encoder layer and the second Controller is located after the sixth encoder layer. We use the same positioning system as in the encoder for the Controllers of the decoder. The last two columns show the \textit{Communication} (\textit{C-Cost}) and model \textit{Training} (\textit{T-Cost}) costs, where W stands for the word embedding table.}
\end{center}
\end{table*}

\subsection{Baseline Results}
Rows 1 to 5 in Table \ref{results} report BLEU \citep{papineni2002bleu} scores obtained from our stand-alone engines. These models are trained in a non-FL setting with their own training sets, e.g. \textit{6E-6D WMT} is a Transformer with 6 encoder and 6 decoder layers trained on the WMT dataset. As expected, the model shows its best performance on in-domain data but it fails to successfully handle other domains' data. The lowest BLEU score belongs to the PHP test set. Clearly, technical terms and special form of sentences in the PHP dataset is not fully understandable for the WMT model and makes it challenging for the model to generate high-quality translations. We see a similar trend for other models too, namely they perform well on in-domain data but could barely work for out-of-domain samples. However, for datasets from similar domains this performance gap is small(er), e.g. the \textit{OS} model is able to handle the \textit{TED} testset relatively well as there are lots of similarities between the \textit{OS} (film subtitles) and \textit{TED} (TED talks) sets. \textit{PHP} is an extreme case in our pool; due to its size and technical genre with unique terms and special grammatical constituents, it cannot consume other domains' sentences, so the BLEU scores are almost zero for all out-of-domain datasets. 

In our setting, a single model trained on in-domain data cannot be a good translation engine to support all domains. Therefore, we train a mixed-domain engine to address this problem. Rows 6 and 7 report related results. In the \textit{Fine Tuning} approach (Row 6), we combined all datasets to create a huge corpus of training and development sets. We then train a translation engine using such a collection which leads to high-quality results. The average BLEU score of this new model is 32.00 (higher than all others), which is a clear indication of its success. This model is capable of handling different domains quite well. However, this approach relies on a strong assumption, i.e. that all datasets are fully accessible at training time in a centralized repository, which might not be the case in real-world scenarios, and especially in distributed and private settings such as FL. Therefore, we propose another alternative, namely \textit{Chained Training} (Row 7), where we start from a well-trained model and gradually fine-tune it with other datasets. 

For \textit{Chained Training}, we assume that the WMT model is our base model. We share it with a client and it can fine-tune it with its own data. After each phase of fine-tuning, the updated model is shared with another client, so that it can also add its own data. In this scenario, there is no central training and the model is gradually updated by different clients. As seen in Row 7, results obtained are not as competitive as those in Row 6. This shows how model training can be dramatically impacted in the presence of non-centralized data repositories. It should be noted that, Row 7 does not even address the privacy aspect of the problem and only tries to gradually access data from different clients. Rows 6 or 7 can be considered as a baseline for our FL model, where Row 7 is a more relevant and fair candidate. The setting defined in Row 6 is only suitable for research purposes which discards complications of real-world settings. 

\subsection{FL Experiments}
Row 8 in Table \ref{results} is our first FL system where we have five clients. Each client’s model is initialized with the WMT engine parameters and is updated by the local data in each round of FL. The client model’s parameters are sent to the server at the end of each round and the server aggregates clients’ information via {\fontfamily{pcr}\selectfont FedAVG}. The average BLEU score of this FL system is 33.83, which was impressive and at the same time unexpected given the fact that the server has no access to clients’ data, and the entire process is carried out in a decentralized and private environment. The communication cost (C-Cost) for each client is ``\textit{6E-6D}'' which means it exchanges 6 encoder and 6 decode layers with the server. The training cost (T-Cost) is ``\textit{6E-6D + W}'' since every client updates all the network parameters including those of encoder and decoder, and word embeddings. 

In Row 9, we introduce a new FL configuration where instead of 6 layers we use 8 (for each encoder and decoder). Controllers define dedicated and extra layers responsible for communication and model updates. This increases the number of layers from 6 to 8, so Row 9 would be a fair baseline for Controller-based models. The expectation for Row 9 was a higher BLEU score than that of Row 8, since we increased the number of layers, and in general the quality of deep learning models should increase proportionally with the increase of their size. The same assumption also holds for our model, %, and in fact we observed the same boost. 
but the larger model in Row 9 requires more steps of training to generate better results. For all these models, we ensure that at the end of the training process, either centralized or distributed, each model is only trained for a total of 150K steps.\footnote{The core model, WMT, is trained for 100K steps then we run extra 50K steps of fine-tuning during FL rounds.} In order to have a fair setting, we also trained the 8E-8D/A-A model for 150K steps. 

In Row 10, we apply the Controller idea for the first time. We freeze all layers during fine-tuning on the client side and only update Controller layers. The \textit{T-Cost} for this setup is ``2E-2D'', which is significantly lower than all previously reported results. We still communicate all the encoder/decoder layers in-between clients and the server, so \textit{C-Cost} is still ``8E-8D''. In this configuration, the first and second Controller layers are placed after the second and sixth encoder layers, respectively. We have an identical setting for the decoder Controllers. In Row 11, we have another configuration with a different positioning strategy. The average BLEU score of latter configuration is higher than the previous one, and it shows how placing Controllers in the right place can impact model quality. In Row 12, we have the same setting as in Row 10, but this time not only do we freeze Controllers during fine-tuning but we also limit the FL communication to these layers. Both \textit{C-Cost} and \textit{T-Cost} in this case is ``2E-2D'' which is considerably more affordable. Each encoder layer has 3,416,320 and each decoder layer has 4,204,032 parameters. The embedding table is a matrix with 33,116,512 parameters. Considering these numbers, the bandwidth used (for both communication and training) in Rows 9 and 12 are worth 94,079,328 and 15,240,704 parameters, respectively, which means that our Controller-based models \textit{are about 6 times less expensive}. Another interesting observation is that we are able to achieve better performance than all other 8E-8D models in Row 12, even though we update/communicate fewer parameters. 

In Rows 14 and 15, we challenged our proposed solution and instead of assigning additional layers we pick some layers of the 6E-6D model to play the role of Controllers. Our findings show that these 6E-6D models can perform as accurately as their 8E-8D peers, if we pick the right layers as the Controllers. However, 6E-6D models are much more sensitive to the position of Controllers.

\subsection{Controllers Vs Adapters}
Adaptors (Section \ref{TL}) follow a similar logic as our Controllers, but they are solutions for centralized settings. In Adapters, a small unit is implanted \textit{inside} each layer of a neural model whose job is (\textit{i}) to connect two consecutive layers to each other in a way that they do not deviate from their main distribution, and (\textit{ii}) take care of the transition between domains. Such a similar architecture persuaded us to try Adapters in our FL setting but our experiments were not successful. Our investigation showed that when an Adapter unit is sent out to the server side to be combined with other Adaptors, it becomes so distant from its original form that when it is placed back in the body of a client model, it destroys the internal information flow and stops the client from converging. We ran preliminary experiments with Adapters instead of Controllers and obtained gravely lower BLEU scores. 

Aside from this qualitative aspect of the problem, we also do not find Adapters suitable for FL settings, because of a design overhead they introduce, e.g. some references recommend that the best position for Adapters to be placed is before Layer Norm \citep{ruckle2020adapterdrop}, others placed them after Feed Forward sub-layers \citep{bapna2019simple}, and the placement options are not limited to only these alternatives. Moreover, each model proposes a different internal neural architecture for Adapters. These design decisions are made through exhaustive empirical investigations. This form of architecture search is not affordable in distributed learning settings like FL. Any exchange of information and re-run of training could be quite costly. 

This design dilemma is not unique to Adapters, and our Controllers could also suffer from similar issues, but in the Controller case the situation is far less severe. As Table \ref{results} shows, we have tried different positioning strategies and our findings show that positioning Controllers layers affect the final quality. This also becomes even more important when working with smaller models (6E-6D vs 8E-8E). However, this does not stop us from training clients in the FL setting, because almost any random positioning strategy leads to \textit{acceptable} results. There is only one exception and that is when we placed the first Controller in between the embedding table and the first layer of the encoder. If we define a configuration such as \textit{8E-8D/A-C (-1-6)} where the first layer of an encoder/decoder is a Controller, a client model would never converge. We could not find any mathematical justification for this observation, but we assume that because the first layers plays a critical role in building input representations, any noisy information can affect the entire model's sustainability. It seems information added from other domains to the first layer after the embedding table, when it is used as a Controller, has a similar effect and stops clients from converging.

\section{Conclusion}\label{conc}
In this work, we focused our effort on developing a communication- and computation-efficient FL solution to train NMT models. Our research proposed a novel idea to include what we call \textit{Controller} layers. These layers are the ones trained, while the original network layers are frozen. By doing so, these layers learn the necessary information and communicate it to the central server during updates. The server aggregates the information from all the clients, and then sends the updates to each client through the Controllers. Our Controller-based models are about six times less expensive to train in comparison to baseline NMT models. This reduction is significantly important when we consider the large number of model parameters that are exchanged during multiple update rounds. For our future work, we are planning to apply Controllers to other tasks and discover a systematic way of finding the best position for placing Controllers.

%%%%%%%%%%%%%%%%%%%%%%%%%%%%%%%%%%%%%%%%%%%%%%%%%%%%%%%%%%%%
\bibliography{neurips_2021}
\bibliographystyle{neurips_2021}
%%%%%%%%%%%%%%%%%%%%%%%%%%%%%%%%%%%%%%%%%%%%%%%%%%%%%%%%%%%%
%%%%%%%%%%%%%%%%%%%%%%%%%%%%%%%%%%%%%%%%%%%%%%%%%%%%%%%%%%%%
%%%%%%%%%%%%%%%%%%%%%%%%%%%%%%%%%%%%%%%%%%%%%%%%%%%%%%%%%%%%

%%%%%%%%%%%%%%%%%%%%%%%%%%%%%%%%%%%%%%%%%%%%%%%%%%%%%%%%%%%%

\end{document}